%% file: egpaper_for_review.tex
\documentclass[10pt,twocolumn,letterpaper]{article}

\usepackage{3dv}
\usepackage{times}
\usepackage{epsfig}
\usepackage{graphicx}
\usepackage{amsmath}
\usepackage{amssymb}
\usepackage{booktabs}
\usepackage{makecell}
\newcommand\minisection[1]{\vspace{1mm}\noindent \textbf{#1}}

\usepackage[pagebackref=true,breaklinks=true,colorlinks,bookmarks=false]{hyperref}

\threedvfinalcopy % *** Uncomment this line for the final submission

\def\threedvPaperID{125} % *** Enter the 3DV Paper ID here

\ifthreedvfinal\fi
\begin{document}

%%%%%%%%% TITLE
\title{Semi-supervised 3D Object Detection via Temporal Graph Neural Networks}

\author{Jianren Wang\\
Carnegie Mellon University\\
{\tt\small jianrenw@andrew.cmu.edu}
% For a paper whose authors are all at the same institution,
% omit the following lines up until the closing ``}''.
% Additional authors and addresses can be added with ``\and'',
% just like the second author.
% To save space, use either the email address or home page, not both
\and
Haiming Gang\\
Honda Research Institute\\
{\tt\small hgang@honda-ri.com}
\and
Siddharth Ancha\\
Carnegie Mellon University\\
{\tt\small sancha@cs.cmu.edu}
\and
Yi-Ting Chen\\
National Yang Ming Chiao Tung University\\
{\tt\small ychen@cs.nctu.edu.tw}\and
David Held\\
Carnegie Mellon University\\
{\tt\small  dheld@andrew.cmu.edu}
}

\maketitle
% \thispagestyle{empty}

%%%%%%%%% ABSTRACT
\begin{abstract}
% \dave{Replace the **** at the top “Paper ID ****” with the actual paper ID.}
3D object detection plays an important role in autonomous driving and other robotics applications. However, these detectors usually require training on large amounts of annotated data that is expensive and time-consuming to collect. Instead, we propose leveraging large amounts of unlabeled point cloud videos by semi-supervised learning of 3D object detectors via temporal graph neural networks. Our insight is that temporal smoothing can create more accurate detection results on unlabeled data, and these smoothed detections can then be used to retrain the detector. We learn to perform this temporal reasoning with a graph neural network, where edges represent the relationship between candidate detections in different time frames. After semi-supervised learning, our method achieves state-of-the-art detection performance on the challenging nuScenes~\cite{caesar2020nuscenes} and H3D~\cite{360LiDARTracking_ICRA_2019} benchmarks, compared to baselines trained on the same amount of labeled data. Project and code are released at \url{https://www.jianrenw.com/SOD-TGNN/}.
\end{abstract}

%%%%%%%%% BODY TEXT
\input{sections/1-introduction.tex}

\input{sections/2-related_works.tex}

\input{sections/3-approach.tex}

\input{sections/4-results.tex}

\input{sections/5-conclusion.tex}

{\small
\bibliographystyle{ieee_fullname}
\bibliography{egbib}
}

\end{document}

%%%%%%%%% TITLE
\title{Semi-supervised 3D Object Detection via Temporal Graph Neural Networks \\ Supplementary Material for 3DV 2021 Paper \#125}

\author{First Author\\
Institution1\\
Institution1 address\\
{\tt\small firstauthor@i1.org}
% For a paper whose authors are all at the same institution,
% omit the following lines up until the closing ``}''.
% Additional authors and addresses can be added with ``\and'',
% just like the second author.
% To save space, use either the email address or home page, not both
\and
Second Author\\
Institution2\\
First line of institution2 address\\
{\tt\small secondauthor@i2.org}
}

\maketitle

\section{Training Details}

\subsection{Graph Augmentation}

While performing data augmentation, we copy and paste trajectories from other videos such that it is at most 10m away from a candidate detection. We randomly copy $m$ trajectories (m is uniformly sampled from 1-5) from other videos and randomly remove $m$ trajectories from the augmented video. We use 4 preceding frames and 4 succeeding frames as neighbouring frames ($N=4$ as denoted in Section 3.1). Combined with the current frame, a node can be connected to 9 frames in total.

\subsection{Teacher: Temporal Graph Neural Network}

The graph neural network operations can be written as:
\begin{equation}
    x^k_i = \gamma^k(x^{k-1}_i, \cup_{j\in\mathcal{N}(i)}\phi^k(x^{k-1}_i,x^{k-1}_j,e_{j,i}))
\label{eq:forward}
\end{equation}
where $\gamma^k$ denotes the $k^{th}$-step updating network, $\phi^k$ denotes the $k^{th}$-step message network, $\cup$ denotes the message aggregation function, $x^{k-1}_i$ and $x^{k-1}_j$ denotes node features of node $i$ and $j$ in layer (k-1), and $e_{j,i}$ denotes the edge features of node $j$ and node $i$.

We use a single fully connected layer for both updating network and message network at each step. We iterate the message aggregation and updating procedure 4 times, which forms a 4-layer-GNN. The total number of hidden neurons (width) is 8. We use the mean function for message aggregation. We use the default initialization in PyTorch to initialize our network, which is a variant of Kaiming He Initialization~\cite{he2015delving}.

The temporal GNN of the teacher is initialized from 50 labeled videos of nuScenes. We use the Adam optimizer~\cite{kingma2015adam} for training the network with a learning rate of $1\times10^{-3}$, $\beta_1$ and $\beta_2$ are 0.9 and 0.999, respectively. We train the network for 50 epochs.

\subsection{Semi-supervised Training}

We use the original CenterPoint \cite{yin2021center} architecture and training procedure, but modify the loss to the new semi-supervised loss. To obtain the initial student model, we train the detector of student model for 20 epochs on the labeled data. During the semi-supervised training, we first obtain the detection results on the unlabeled data from the detector of student, and then use GNN of student to modify the detection score according to temporal information and histogram binning to calibrate the score. Finally, we apply the Equation 6 on calibrated score to estimate the uncertainty of each object. We use the mixed labeled and pseudo labeled data to train the detector and GNN of the student individually for 5 epochs as one iteration, and then use the same pseudo label generation process to obtain the new pseudo labels for next iteration training. The models were trained on a single NVIDIA Quadro V100 GPU.

\subsection{Flicker}

For each individual detection, we first calculate the average confidence $C_\text{avg}$ of the ``nearby" detections (where ``nearby" is defined according to the distance between the forecasted position using velocity and the actual position i.e. motion projection of Equation 1) in each adjacent frames. The detection is then rescored to $s_\text{new} = C_\text{avg} \cdot s_\text{old}$, where $s_\text{old}$ is the original detection score. Specifically, we first project an object to its neighboring frames based on its estimated velocity using Equation 1. Detections whose distance to the predicted position of the object is less than 10 meters are considered as ``nearby" detections. We consider 4 preceding frames and 4 succeeding frames as adjacent frames.

\section{Visualization}

Please see the attached video for more visualizations.

\section{Code}
Please refer to the attached code to see our implementation.

{\small
\bibliographystyle{ieee_fullname}
\bibliography{egbib}
}

%% file: sections/1-introduction.tex
\section{Introduction}

3D object detection is an essential and fundamental problem in many robotics applications, such as autonomous driving~\cite{chen2020pano3d}, object manipulation~\cite{weng2020multi}, and augmented reality~\cite{10.1007/978-3-319-46484-8_10}. In recent years, many deep learning-based approaches for point cloud-based 3D object detection~\cite{yan2018second, lang2019pointpillars, zhu2019class} have emerged and achieved high performances on various benchmark datasets~\cite{sun2020scalability,caesar2020nuscenes,Geiger2013IJRR,360LiDARTracking_ICRA_2019}. Despite the impressive performances, most of the existing deep learning-based approaches for 3D object detection on point clouds are strongly supervised and require the availability of a large amount of well-annotated 3D data that is often time-consuming and expensive to collect.

\begin{figure}
    \centering
    \includegraphics[width=0.9\linewidth]{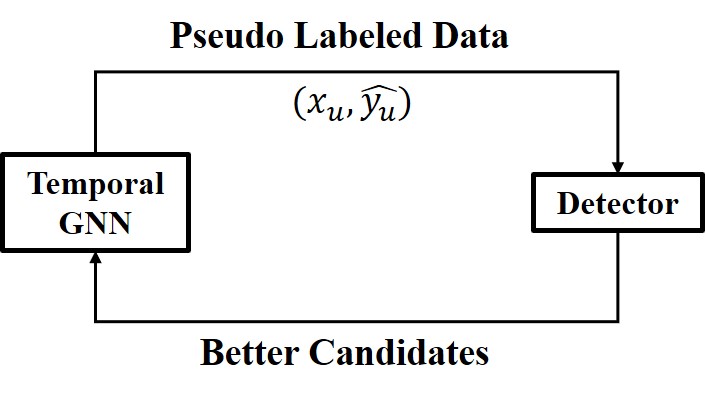}
    \caption{\textbf{Semi-supervised 3D Object Detection via Temporal Graph Neural Networks.} 
    % Given a large amount of unlabeled point cloud videos, we build video graphs,
    % where each node represents a candidate detection predicted by a pretrained detector and each edge represents the relationship between the connected nodes in different time frames. \dave{You are jumping into how without first explaining why; I would probably keep this caption at a higher level to differentiate it from the caption of Fig 3} A graph neural network is then used to refine these candidate detections using  spatiotemporal information. These  pseudo labeled point clouds are then combined with labeled point clouds to retrain the detector. This detector can be used to generate better candidates, which will in turn lead to better psuedolabels.  \dave{Make sure to refer to this figure at some point in the paper}
    Our method utilizes the rich spatiotemporal information from point cloud videos to perform semi-supervised learning to train a single frame  object detector. This detector can be used to generate better candidates, which will in turn lead to better psuedo labels.
    }
    \label{fig:teaser}
\end{figure}

Semi-supervised learning~\cite{teichman2012tracking} is a promising alternative to supervised learning for point cloud-based 3D object detection. This is because semi-supervised learning requires only a limited amount of labeled data, instead relying on large amounts of unlabeled data to improve performance. The challenge with semi-supervised learning is to determine how to make use of the unlabeled data to improve the performance of the detector.

% Furthermore, the available few strong labels can still provide the necessary supervision to guide the deep network into learning the correct information for 3D object detection. Zhao et al.~\cite{Zhao_2020_CVPR} achieves semi-supervision with a Mean Teacher paradigm~\cite{tarvainen2017mean} that contains a teacher and student 3D object detection network. The teacher guides the predictions of the student to be consistent with its predictions under random perturbations, where these predictions are sets of 3D object proposals. Thanks to the abundance of pseudo labeled data and the use of random perturbations, the student learns to become better than the teacher. Despite the strong performance of Pseudo Labels methods, they have one main drawback: if the pseudo labels are inaccurate, the student will learn from inaccurate data. To correct the confirmation bias, Pham et al.~\cite{pham2020meta} utilizes the feedback from the student to inform the teacher to generate better pseudo labels. However, this method still cannot correct the inaccurate predictions. 

\begin{figure}
    \centering
    \includegraphics[width=0.9\linewidth]{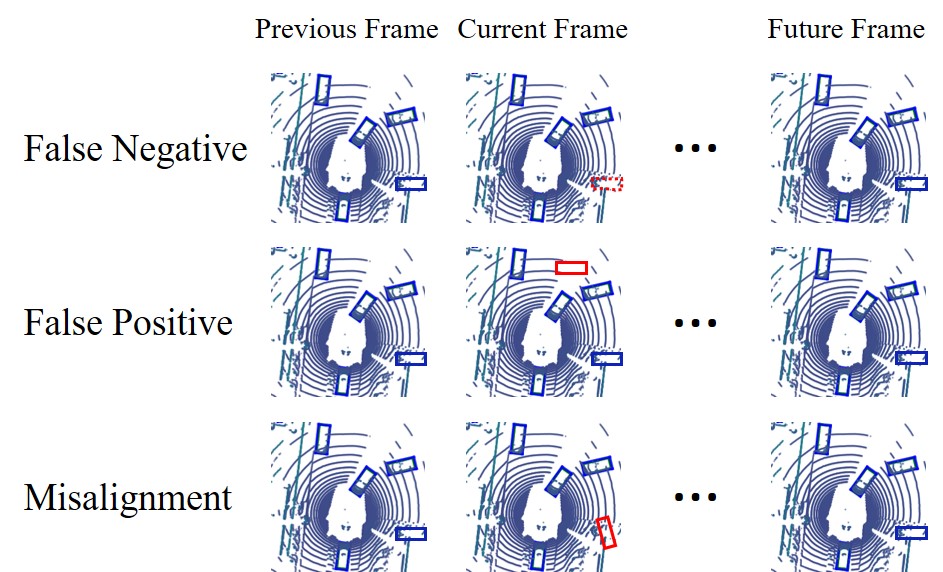}
    \caption{\textbf{Detections in Point Cloud Videos.} 
    % \dave{This image is coming without context; explain that this image illustrates common failure cases of object detectors; then explain how these failures can be detected using spatiotemporal reasoning} 
    3D object detectors usually suffer from false negatives, false positives and misalignment due to the sparse nature of LiDAR data. However, these failures can always be detected by exploiting the rich spatiotemporal information in point cloud videos. For example, a false negative  (red box in the top row) can be identified if the same object is detected in adjacent frames but is only missing in the current frame. A false positive (red box in the middle row) can be identified if the detection isisolated i.e. a corresponding detection occurs neither in the previous nor the subsequent frame. A misalignment (red box in the bottom row) can be identified if the same object is detected in adjacent frames but the alignment differs significantly in the current frame. (Blue boxes show true positives, where the same object is detected in all adjacent frames).}
    % Blue boxes show true positives, where the same object is detected in all adjacent frames. In the top row, the red box shows a false negative (\textit{i.e.} the same object is detected in adjacent frames but only missing in the current frame). In the middle row, the red box shows a false positive (\textit{i.e.} no object exist in adjacent frames but an object is falsely detected in the current frame). On the bottom row, the red box shows a misalignment (\textit{i.e.} the same object is detected in adjacent frames but the alignment differs significantly in the current frame.)}
    \label{fig:pc_video}
\end{figure}

In most applications, point clouds  are recorded over time as a data stream. A point cloud video contains richer spatiotemporal information than a single frame.  Our insight is that this  spatiotemporal information can be exploited to correct  inaccurate predictions. For example, a false negative (missing) detection can be identified if the same object is detected in adjacent frames but is missing in the current frame. A false positive detection can be identified if the detection is isolated \textit{i.e.} a corresponding detection occurs neither in the previous nor the subsequent frame. A misalignment can be identified if the alignment differs significantly between successive video frames (see Figure~\ref{fig:pc_video} for examples).

In this paper, we propose utilizing the rich spatiotemporal information from point cloud videos to perform semi-supervised learning to train a single frame object detector (Figure~\ref{fig:teaser}).
Specifically, we first build video graphs using a large amount of unlabeled point cloud videos, where each node represents a candidate detection predicted by a pretrained detector and each edge represents the relationship between the connected nodes in different time frames. A graph neural network is then used to refine these candidate detections using spatiotemporal information. These refined detections are treated as pseudo labels and are used to re-train a detector. This retrained detector can be used to generate better detections, which can again be refined in an iterative process of continual improvement. 

%Thus, the temporal graph neural network and detector can be trained iteratively, which will lead to better and better object detector.

A potential limitation with learning using pseudo labels generated from temporally smoothed detections is that these labels are not necessarily accurate. The detector may produce mistakes, and learning on these mistakes may lead to inferior 3D detections. We propose to explicitly tackle this problem by incorporating uncertainty into semi-supervised learning. Specifically, we propose to use the entropy of estimated detections as uncertainty weights for the semi-supervised loss. We demonstrate the efficacy of our method across a variety of large-scale benchmark datasets, including nuScenes~\cite{caesar2020nuscenes} and H3D~\cite{360LiDARTracking_ICRA_2019} and show state-of-the-art detection performance, compared to baselines trained on the same amount of labeled data.

%Our method is plug-and-play and can be applied to any 3D point cloud video datasets.

The contributions of this paper are as follows:
\begin{enumerate}
    \item We propose a novel framework for semi-supervised 3D object detection by leveraging the rich spatiotemporal information in 3D point cloud videos.
    \item We propose a \textit{3DVideoGraph}  for spatiotemporal reasoning in 3D point cloud videos.
    \item We show that we can use the 3DVideoGraph with uncertainty loss weighting for semi-supervised training of 3D object detectors.
    \item We demonstrate our method over two large-scale benchmark datasets and show state-of-the-art detection performance, compared to baselines trained on the same amount of labeled data.
\end{enumerate}

%% file: sections/2-related_works.tex
\section{Related Work}

\paragraph{3D Object Detection}
3D object detectors aim to predict 3D oriented bounding boxes around each object. A common approach to 3D object detection is to exploit the ideas that have been successful for 2D object detection~\cite{girshick2014rich, girshick2015fast, ren2015faster, he2017mask}, which first find category-agnostic bounding box candidates, then classify and refine them. Existing works can be roughly categorized into three groups, which are birds-eye-view based methods~\cite{yang2018pixor, liang2018deep}, voxel-based methods~\cite{zhou2018voxelnet, yan2018second}, and point-based methods~\cite{lang2019pointpillars,vora2020pointpainting}. Unlike these methods, our work focus on semi-supervised learning to improve a detector's performance.

\paragraph{3D Video Object Detection}
A few recent papers have incorporated spatiotemporal reasoning in 3D video object detection. 3DVID~\cite{yin2020lidar} proposes an Attentive Spatiotemporal Transformer GRU (AST-GRU) to aggregate spatiotemporal information across time. Similarly, 3DLSTM~\cite{huang2020lstm} proposes a sparse LSTM to aggregate features across time. These works show great potential by utilizing the rich spatiotemporal information in point cloud videos. However, they use memory-intensive sequence models to utilize the temporal information; as a result, only three consecutive frames can be input to the model due to memory limitations~\cite{yin2020lidar}, which makes it hard to reason about complex spatiotemporal information. In contrast, our graph neural network representation can reason over much longer sequences. Further, we show how such spatiotemporal reasoning can be combined with uncertainty estimates for semi-supervised learning of a single-frame detector.

\paragraph{Semi-supervised Learning}
Many approaches have been proposed for semi-supervised learning (SSL), which learns from a small labeled dataset combined with a much larger unlabeled dataset. One approach uses pseudo labels, also known as self-training. Self-training has been successfully applied to improve the state-of-the-art of many tasks, such as image classification~\cite{xie2020self}, object detection~\cite{liu2021unbiased}, semantic segmentation~\cite{sun2020teacher}. These methods often involve a teacher which provides pseudo-labels for a student which learns from these pseudo-labels~\cite{arazo2020pseudo,pham2020meta}. However, these methods highly depend on the performance of teacher. which often makes incorrect predictions. 

Another promising direction for SSL is self-ensembling, which encourages consensus among ensemble predictions of unknown samples under small perturbations of inputs or network parameters~\cite{Zhao_2020_CVPR,miyato2018virtual, wang20213dioumatch}. The student learns to perform better than the teacher due to its robustness to corruption. However, the improvement is limited since the teacher and student can use the same data to make predictions. In contrast, we propose to use spatiotemporal information to construct a better teacher, which is then used to train a student which has access to only single-frame information. 

\paragraph{Spatiotemporal Reasoning}
Most efforts in spatio-temporal reasoning focus on 2D semantic segmentation~\cite{yang2019step, Nilsson_2018_CVPR, perazzi2017learning}. For example, Bao et al.~\cite{bao2018cnn} embed mask propagation into the inference of a spatiotemporal MRF model to improve
temporal coherency. EGMN~\cite{lu2020video} employs an episodic memory network to store frames as nodes and capture cross-frame correlations by edges. However, these methods are computationally expensive even in 2D videos, which makes them infeasible to be adapted to 3D videos. In contrast, our temporal GNN is very computational efficient and memory efficient and thus can be applied to long sequences.

%% file: sections/3-approach.tex
\section{Approach}

\begin{figure*}
    \centering
  \includegraphics[width=0.7\textwidth]{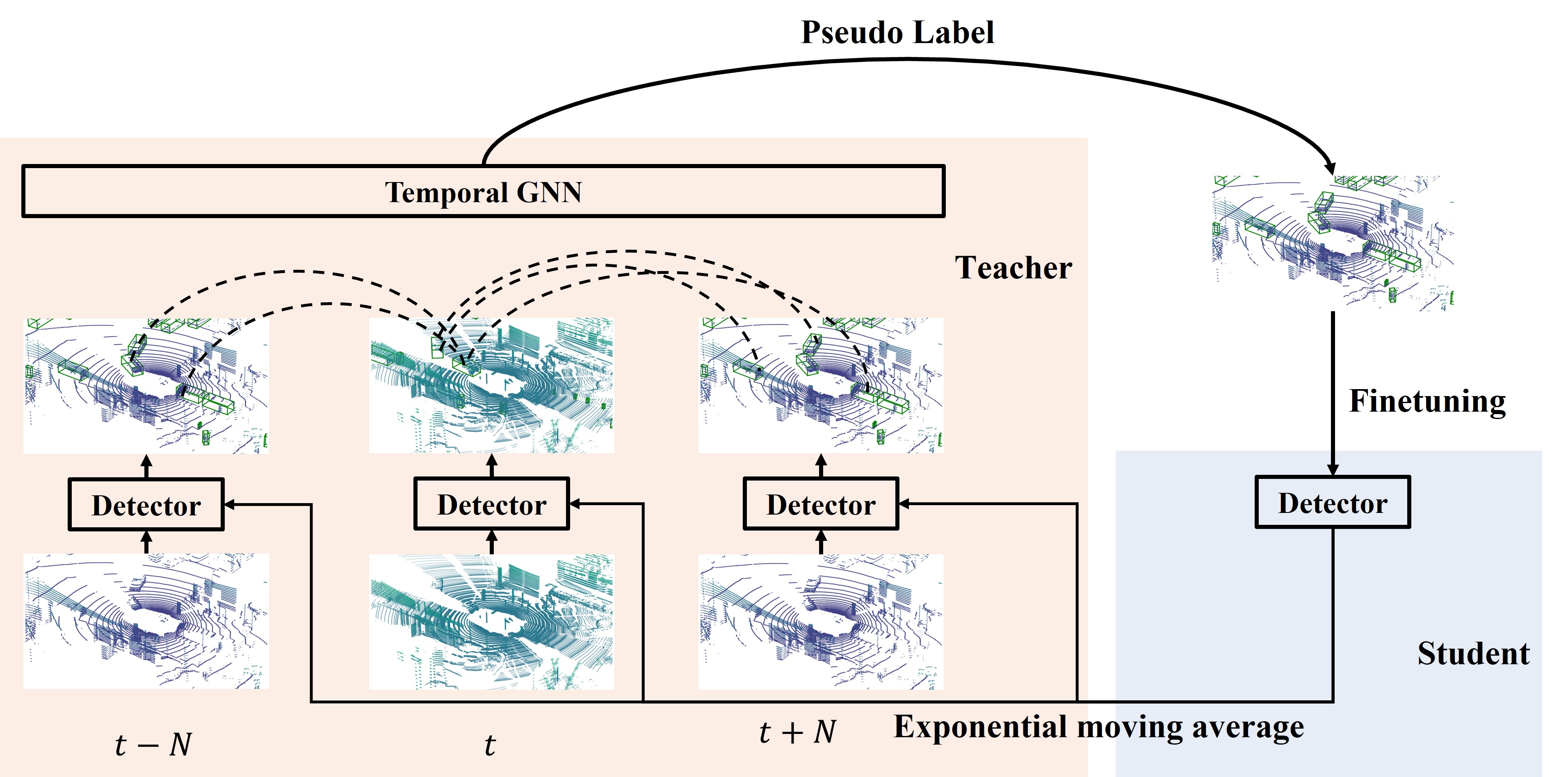}
  \caption{Overview of Semi-supervised 3D Object Detection via Temporal Graph Neural Networks: The teacher consists of a 3D object detector and a graph neural network for spatiotemporal reasoning. The 3D object detector takes a single point cloud frame as input and outputs candidate detections. The graph neural network takes candidate detections from a sequence of point clouds as input and outputs refined detection scores. These pseudo labeled point clouds are then combined with labeled point clouds to train the student. 
%   This detector can be  used  to  generate better  candidates,  which  will  in  turn  lead  to better pseudo labels. Thus, the temporal graph neural network and detector can be trained iteratively, which will lead to better and better object detector.
The 3D object detection module of the teacher and the student are with the same architecture. And the parameters of the 3D object detection module of the teacher are the exponential moving average of the continually updated student network parameters.  This updated detector is used to generate better detections, which is further used to refine the spatiotemporal reasoning module.  The iterative refinement of student and teacher can lead to a continual improvement.}
  \label{fig:pipeline}
\end{figure*}

In this section, we elaborate on our method for semi-supervised 3D object detection. Our method consists of a teacher which performs spatiotemporal reasoning for 3D object detection; this teacher is then used to provide pseudo labels to train a student which takes as input only a single frame (see Figure~\ref{fig:pipeline}). We use uncertainty-aware training to handle incorrect pseudo labels produced by the teacher.

\subsection{Teacher Network: GNN for Spatiotemporal Reasoning}
\label{sec:teacher}
We first describe our teacher network, which uses spatiotemporal reasoning to provide pseudo labels on unlabeled data. These pseudo labels are then used to train a student which takes as input single-frame data. 

% \dave{Rephrase this; our teacher is a 3D object detector; initially trained from X and later retrained from Y} 

Our teacher network consists of two modules: a 3D object detection module, and a spatiotemporal reasoning module. The 3D object detection module can be any existing 3D object detector, which is initially trained from only a limited amount of labeled data. We then input each frame to the 3D object detector to find detected objects in the scene. We filter the output of this detector with a fairly low threshold (we use a confidence of 0.1 in our experiments) to create a large set of candidate detections in the scene. However, these predictions are always noisy and inaccurate. 

To smooth these detections and create more accurate pseudo labels, we propose a novel spatiotemporal reasoning module: we build a video graph over the frames of the 3D video (sequence of point clouds), and use a Graph Neural Network (GNN) to output new scores for each detection node based on the principle of spatiotemporal consistency. This spatiotemporal reasoning module is trained on the same labeled data as the 3D object detection module.

\label{sec:feature_extractor}
Specifically, we extract a set of high-level features of each candidate detection as node features ($x$) for the 3D video graph. These high level features include (1) the detection score, (2) the number of points in the detection box, and (3) the size of the detection box (width, length, and height). In total, our node features are 5-dimensional. 

% Our teacher can be built on top of any existing single-frame 3D object detector. We input each frame to the 3D detector to find detected objects in the scene. We filter the output of this detector with a fairly low threshold (we use a confidence of 0.1 in our experiments) to create a large set of candidate detections in the scene. For each candidate detection, we extract a set of high-level features which are used as node features for the 3D video graph. These high level features include (1) the detection score, (2) the number of points in the detection box, and (3) the size of the detection box (width, length, and height). In total, our node features are 5-dimensional. \dave{Use a variable for the node features}. After feature extraction, we have a node feature vector for each candidate detection in each frame of a video. 

%\subsection{GNN for Spatiotemporal Reasoning}
% We then construct 
% a graph to perform spatiotemporal reasoning \dave{why?}.
% \dave{Move to later: we construct a separate graph for each category.} 
% To connect nodes across frames, we use the detection's estimated velocity, which is often predicted by the 3D object detector~\cite{zhu2019class, chen2020pano3d, yin2021center}.  If the 3D object detector does not estimate the object's velocity, we assume zero velocity, although tracking methods could also be used here.
% \dave{Too much explaining ``how" without explaining ``why" - I don't understand what we're trying to achieve here}

\label{sec:gnn_reasoning}
After feature extraction, we have a node feature vector for each candidate detection in each frame of a video. To connect nodes across frames, we use the detection's estimated velocity, which is often predicted by the 3D object detector~\cite{zhu2019class, chen2020pano3d, yin2021center}. We project the node to its neighboring frames based on its estimated velocity, as follows:
\begin{equation}
    \hat{p}^i_{t+N} = v^i_t \cdot (T_t - T_{t+N}) + p^i_t,
\label{eq:forward}
\end{equation} where $\hat{p}^i_{t+N}$ represents the predicted position of object $i$ in frame $t+N$, $p^i_t$ and $v^i_t$ represents the estimated position and velocity of object $i$ in frame $t$, and $T_t$ and $T_{t+N}$ represents the timestamp of frame $t$ and frame $t+N$. We then calculate the distance between the predicted position of object $i$ ($\hat{p}^i_{t+N}$) with all detections in frame $t+N$. All detections whose distance to the predicted position are smaller than a threshold are connected to node $i$. Each node is connected to all such neighboring nodes in the neighboring frames from $t-N$ and $t+N$. We also add the following edge features ($e$) to the 3D video graph: (1) the distance between the predicted center positions of each pair of detection boxes (1-dimensional), (2) the difference in the sizes along each dimension (3-dimensional), and (3) the difference in bounding box orientations (1-dimensional). In total, our edge features are 5-dimensional.

% \dave{The below paragraph should probably be moved to the second paragraph of this section}
% Finally, we perform inference over this graph to output new scores for each detection node. \dave{It should be stated much earlier that this is what we are trying to achieve with the GNN} Specifically, 

Finally, our GNN takes the node feature and edge feature as input, and it outputs a new detection score $s_i$ for the $i^{th}$ candidate detection. The graph neural network operations can be written as:
\begin{equation}
    x^k_i = \gamma^k(x^{k-1}_i, \cup_{j\in\mathcal{N}(i)}\phi^k(x^{k-1}_i,x^{k-1}_j,e_{j,i})),
\label{eq:inference}
\end{equation} where $\gamma^k$ denotes the $k^{th}$-step updating network, $\phi^k$ denotes the $k^{th}$-step message network ($k^{th}$-layer of the graph neural network), $\cup$ denotes the message aggregation function, $x^{k-1}_i$ and $x^{k-1}_j$ denotes node features of node $i$ and $j$ in layer $(k-1)$, and $e_{j,i}$ denotes the edge features of node $j$ and node $i$, which remains unchanged for all layers. The output of the final update is the new detection score $s_i$.

% \dave{Write this as an equation}

\begin{equation}
    s_i = x_i^K,
\label{eq:score}
\end{equation} where $x_i^K$ represents the final update (output of the final layer) of node $i$. 

The GNN is trained end-to-end using the Binary Cross Entropy Loss (BCE). The spatio-temporal reasoning of the GNN allows the network to output more accurate detections than those of the original 3D object detector. 
% \dave{Section missing} for more details on the training procedure.

These refined detections are treated as pseudo labels and are used to train a student (a 3D object detector). For simplicity, the 3D object detection module of the teacher and the student are with the same architecture. And the parameters of the 3D object detection module of the teacher $\theta^T$ are the exponential moving average (EMA) of the continually updated student network parameters $\theta^S$. 

\begin{equation}
    \theta_t^T = \alpha \theta_t^S + (1-\alpha) \theta_{t-1}^T,
\label{eq:weight_update}
\end{equation} where the coefficient $\alpha$ represents the degree of weighting decrease, $\theta_t^S$ represents the student network parameters after iteration $t$. 

This updated detector is used to generate better detections, which is further used to refine the spatiotemporal reasoning module. The iterative refinement of student and teacher can lead to a continual improvement. Please refer to Section~\ref{sec:training_details} for more details on the training procedure.

\subsection{Graph Augmentation}
To prevent the graph neural network from overfitting, we use data augmentation. It's worth noticing all these augmentations are only applied to the limited amount of labeled data since the spatiotemporal reasoning module is only trained on the labeled data. We propose three novel data augmentation techniques.
% \dave{Are you sure that these are novel?  A lit review might be helpful to make sure}:

\paragraph{Copy-and-Paste}

A copy-and-paste augmentation scheme is widely used in many popular single-frame object detectors including SECOND~\cite{yan2018second}, PointPillars~\cite{lang2019pointpillars}, and CenterPoint~\cite{yin2021center}, which crops ground truth bounding boxes from other frames and pastes them onto the current frame's ground plane. Unlike previous methods, we propose trajectory-level copy and paste. We first assign each object an identity based on the its center distance to the closest ground truth object
% \dave{Is this only for the labeled data?  What do you do for the unlabeled data where we don't have ground-truth? We should clarify which parts of our method use labeled data and which parts use unlabeled data}. 
Then, detections with same identity form a trajectory; a random clip of the trajectory is copied into a new video with a random starting frame.
% part of the trajectory or the entire trajectory \dave{This sounds vague} is copied into a new video with a random starting frame. 
% To improve the message passing rate between nodes \dave{Not sure what the ``message passing rate" is}, we  place the starting point of a trajectory only at positions for which the distance to a candidate detection is smaller than a threshold.
% To improve the message passing rate between nodes, we place the starting point of a trajectory only at positions for which the distance to a candidate detection is smaller than a threshold.
% Experimentally, this threshold is the same as used for node connection as mentioned in Section~\ref{sec:gnn_reasoning} \dave{Use a variable for this}. 
Similarly, we also randomly remove trajectories from current video as a form of data augmentation.

\paragraph{Random Trim}
To further increase the diversity of our data, we propose to randomly select the first frame and the last frame from a video to form the sequence of frames used in the graph. 
% \dave{This is too vague; add more details}

\paragraph{Random Noise}
    
We also propose to add random noise to the location, size, orientation and detection scores of individual detections. 
% This is to simulate the diversity of outputs from a pre-trained object detector \dave{What does this mean?  I thought we are training an object detector; why do we need to simulate one?}. 
This is to simulate errors of a pre-trained object detector on more diverse unlabelled data.
The noise values for the location and size are uniformly sampled from $\pm 10\%$ of the bounding box size, while the orientation noise is uniformly sampled from $\pm 10^{\circ}$. The detection score noise is uniformly sampled from $\pm0.15$ and is only applied to detections in the middle of a trajectory (not detections at the beginning or end of a trajectory).

\subsection{Uncertainty-aware Semi-supervised Training}

One of the most prominent challenges in training on pseudo labels is that they are not guaranteed to be accurate. Our solution is to leverage \textit{uncertainty} in the pseudo labels. For examples with high uncertainty, we discount the contribution of the corresponding example during training. This is intended to reduce the effect of incorrect labels in semi-supervised training.

\label{sec:calibration}
We propose a method for estimating the amount of uncertainty of each pseudo label during training time. To estimate uncertainty, we first obtain the new detection score $s_i$ from the temporal GNN mentioned in Section~\ref{sec:gnn_reasoning}. However, these scores are not always well calibrated. In other words, the probability associated with the predicted label cannot reflect its ground truth correctness likelihood. To calibrate the prediction score, we adopt histogram binning~\cite{zadrozny2001obtaining}, 
% arguably the most common non-parametric calibration method \dave{The popularity is not relevant and should not be mentioned here},
a simple non-parametric calibration method.

% which learns a piecewise constant function $f$ to transform uncalibrated outputs $s_i$: $\hat{s_i} = f(s_i)$ where the $\hat{s_i}$ is calibrated detection score.
In short, all uncalibrated predictions $s_i$ are divided into mutually exclusive bins $B_1$,..., $B_M$. Each bin is assigned a calibrated score $\theta_m$; \textit{i.e.} if $s_i$ is assigned to bin $B_m$, then the calibrated score $\hat{s_i} = \theta_m$. More precisely, for a suitably
chosen M, we first define bin boundaries $0=a_1\leq a_2 \leq ... \leq a_{M+1}=1$, where the bin $B_M$ is defined by the interval $(a_m, a_{m+1}]$. The prediction $\theta_i$ are chosen to minimize the bin-wise squared loss:
\begin{equation}
    \mathop{min} \limits_{\{\theta_i\}} \sum_{m=1}^M \sum_{i=1}^n 1(a_m \leq s_i \leq a_{m+1})(\theta_m-s_i)^2,
\label{eq:bin}
\end{equation} where $1$ is the indicator function. Given fixed bins boundaries, the solution to Eq.~\ref{eq:bin} results in $\theta_m$ that correspond to the average number of positive-class samples in bin $B_m$.

% Specifically, for each region of the input \dave{Mention that we bin the input into regions}, isotonic regression learns a function $f$ to minimize the square loss $\sum_{i=1}^N(f(s_i)-y_i)^2$, where $y_i$ is the ground truth binary label of each detection candidate. Because $f$ is constrained to be piecewise constant, we can write the optimization problem as:
% \begin{equation}
%     \mathop{min} \limits_{M,\theta,a} \sum_{m=1}^M \sum_{i=1}^n 1(a_m \leq s_i \leq a_{m+1})(\theta_m-y_i)^2
% \end{equation}, \dave{Put commas inside the equation; also below} subject to $0=a_1\leq a_2 \leq ... \leq a_{M+1}=1$, $\theta_1 \leq \theta_2 \leq ... \leq \theta_M$, where $M$ is the number of intervals; $a_1, ..., a_{M+1}$ are the interval boundaries; and $\theta_1 ... \theta_M$ are the function values \dave{Explain that $\theta_i$ are the function values for region $i$ (we need to first define the regions)}\dave{Explain how $\theta$ and $f$ are related}. Under this parameterization, isotonic regression is a strict generalization of histogram binning in which the bin boundaries and bin predictions are jointly optimized \dave{Did you implement this yourself or use a library for it?}. 
We then use the entropy~\cite{shannon2001mathematical} of the calibrated score $\hat{s_i}$ as a measure of uncertainty $u_i$:
\begin{equation}
    u_i = -\hat{s}_i \log(\hat{s}_i) - (1-\hat{s}_i) \log(1-\hat{s}_i)
\end{equation}

This uncertainty is then used to weight the pseudo labels during training the student -- any existing 3D object detector, which is always composed of a classification branch and a regression branch. 

We first apply this uncertainty to the classification branch:
\begin{equation}
loss_{c}' = \left\{
\begin{array}{lr} 
-(1-u_i)^k \log (p_i), &\textrm{if}\, \hat{s}_i > 0.5\\
-(1-u_i)^k \log (1-p_i), &\textrm{if}\, \hat{s}_i < 0.5\\
\end{array}
\right.
\label{eq:clss_loss}
\end{equation} where $k$ is the focusing parameter $k \geq 0$ which helps the model focuses on the samples with low uncertainty and $p_i$ is the prediction of our student neural network.
% \dave{Teacher network? Or the GNN?}.

We also apply the uncertainty to the bounding box regression branch:
\begin{equation}
loss_{r}' = u_i \times \sum_{b \in (x,y,z,w,,l,h,\theta)} Dis (\Delta b)
\end{equation}
where $\Delta b$ defines the regression residuals between psuedo-labels and student's prediction, and $Dis$ defines the distance metric (\textit{e.g.} Smooth L1 loss).
% \dave{What are these losses used for?  I assume that they are used to train the student but this is not explicitly mentioned}

In total, the semi-supervised loss $loss_s$ is defined as a combination of uncertainty-weighted classification loss and regression loss:
\begin{equation}
    loss_s = loss_{c}' + loss_{r}'
\end{equation}

\subsection{Gradual Semi-supervised Training and Iterative Refinement}

To avoid the student learning from large amount of unreliable pseudo labels, we propose gradual semi-supervised training inspired by~\cite{kumar2020understanding, teichman2012tracking, hong2020learning}. In a nutshell, the student is training with a mix of labelled and unlabelled data, while the amount of unlabelled data increase gradually in each iteration. After each iteration, we update the 3D object detection module of the teacher as a exponential moving average of the student. This updated detector is used to generate better detections. We then retrain the GNN on labeled data for better spatiotemporal reasoning. As the teacher keeps improving, the student can learn from a larger amount of more reliable pseudo labels each iteration. Thus, combining gradual semi-supervised training with iterative refinement of student and teacher can lead to a continual improvement.

% Since 3D point cloud videos always contains more spatiotemporal information than a single point cloud frame, we can train the object detector and temporal GNN iteratively and expect increasingly better performance till converge. \dave{Did we do this?  If so, we should have a plot in the figure showing the performance getting better with each iteration. Also we should clarify after how many iterations we report the results in the table}

%% file: sections/4-results.tex
\section{Experiments}

We evaluate our approach on 3D object detection on the nuScenes dataset~\cite{caesar2020nuscenes} and Honda 3D dataset (H3D)~\cite{360LiDARTracking_ICRA_2019} , which provide a series of sequence of 3D lidar pointcloud with annotated 3D bounding boxes. We also verify the effectiveness of each component of our method by performing an ablation analysis. 

\subsection{Dataset and Experiment Setup}

To obtain enough unlabeled data for semi-supervised training, we re-split the nuScenes dataset as following. We use 50 scenes from nuScenes train for supervised training and 500 scenes from nuScenes train for semi-supervised training. For semi-supervised training, instead of using the ground truth labels, we use the pseudo labels generated by our proposed teacher network. We also use 150 scenes from nuScenes train as validation set. We pick the iteration with best performance on our validation set and reports its performance on 150 videos from nuScenes validation.

For H3D dataset, we use 50 scenes from H3D train for supervised training and 300 scenes from HRI Driving Dataset (HDD) for semi-supervised training, since H3D and HDD datasets have the same data distribution. We pick the iteration with best performance on 30 scenes from H3D validation and reports its performance on 80 videos from H3D test. We compare our methods with baselines using the official metric: mean Average Precision (mAP).

\subsection{Training Details}
\label{sec:training_details}
Our teacher is composed of two modules: 3D object detection module and spatiotemporal reasoning module. We choose CenterPoint~\cite{yin2021center}, a state-of-art 3D object detector as the detection module for both the teacher and the student. We use a 4-layer-GNN for the spatiotemporal reasoning and use the mean function for message aggregation (Eq~\ref{eq:inference}). Node features and edge features are computed as described in Section~\ref{sec:teacher}. The distance threshold below which two detections in adjacent frames are connected is set to 10m. We use 4 preceding frames and 4 succeeding frames as adjacent frames (N = 4 as denoted in Section~\ref{sec:teacher}). Combined with the current frame, a node can be connected to 9 frames in total.

% we apply a single fully connected layer for both updating network and message network at each step. We iterate the message aggregation and updating procedure 4 times, which forms a . The total number of hidden neurons (depth) is 8. We use the mean function for message aggregation. While training the GNN model, the distance threshold (under motion projection: Eq.~\ref{eq:forward}) which two detections in adjacent frames are connected is set to 10m. While performing data augmentation, we copy and paste trajectories from other videos such that it is at most 10m away from a candidate detection. We use 4 preceding frames and 4 succeeding frames as adjacent frames ($N=4$ as denoted in Section~\ref{sec:gnn_reasoning}). Combined with the current frame, a node can be connected to 9 frames in total.

The student network is initialized from the 3D object detection module of the teacher trained on a small amount of labeled data. During semi-supervised training, we add $20\%$ unlabeled data at each iteration whose pseudo labeled is generated by our proposed teacher. The student is trained with a mix of labeled and unlabeled data at each iteration (\textit{i.e.} supervised loss on labeled data and uncertainty-aware semi-supervised loss on unlabeled data). We use the Adam optimizer~\cite{kingma2015adam} for training the student with a batch size of 4 and a learning rate of $1\times10^{-3}$. We also use the Adam optimizer~\cite{kingma2015adam} for training the temporal GNN of the teacher with a batch size of 50 videos and a learning rate of $1\times10^{-3}$. We train the student for 5 epochs and temporal GNN for 5 epochs each iteration. 

% network to generate the pseudo labels. Following the calibration approach mentioned in ~\ref{sec:calibration} to modify the detection scores. Finally, the mixed pseudo and ground truth labels feed into the student network to update the weights of detector. Specifically, we add the $20\%$ of the total unlabeled data for each iteration instead of the whole unlabeled data to help the student network learns gradually which is inspired by \cite{hong2020learning}. 

% We train our method on 700 videos from nuScenes training, and test our method on 150 videos from nuScenes validation. We use the Adam optimizer~\cite{kingma2014adam} for training the network with a batch size of 64 videos and a learning rate of $1\times10^{-3}$. We train the network for 200 epochs and report performance on the test set.
% \label{sec:training_details}

% \dave{This should probably be moved to the end of the approach section} 

\subsection{Results}

We compare our method to the following baselines:

\begin{itemize}
    \item \textit{Student (w/o Semi-supervised Training)}: We provide the performance of the original student initialized from the 3D object detection module of the teacher trained on a small amount of labeled data. 
    \item \textit{Gradual Semi-supervised Training~\cite{kumar2020understanding, teichman2012tracking, hong2020learning}}: In gradual semi-supervised training, the teacher and student share the same architecture (i.e. CenterPoint~\cite{yin2021center}), while the student is trained with a mix of labeled and unlabeled data. The amount of unlabeled data increase gradually in each iteration. Detections with the maximum predicted probability are used as the pseudo label for each unlabeled sample. 
    % \dave{This needs more details; is there a teacher and student network?}
    \item \textit{SESS~\cite{Zhao_2020_CVPR}}: SESS is a self-ensembling semi-supervised 3D object detection framework. During training, labeled samples and unlabeled samples are perturbed and then input into the student and the teacher network, respectively. The student is trained with a supervised loss on labeled samples and a consistency loss with the teacher predictions using on unlabeled samples.
    \item \textit{Oracle (Fully Supervised)}: 
    To provide an upper bound on our performance, we also compare against using full supervision, i.e. the student trained on the same data points (labeled and unlabeled point cloud videos) as semi-supervised training, but is provided ground truth label of all data. This is the ideal case for semi-supervised training and achieves the oracle performance of our method.
\end{itemize}

 For nuScenes, our method performs consistently better than each of the baselines in all categories (Table~\ref{tab:nuScenes}). For H3D, our method outperforms all baseline methods in the overall performance (Table~\ref{tab:h3d}). However, some baselines outperform our method on “other vehicle”. The class of “other vehicle” includes different types of cars and is diverse, with only a small number of examples per vehicle type. Our method also performs marginally worse (within noise) than the baselines on Car and Pedestrian categories. We generally find that classes which have more labeled data (car, pedestrian) benefit less from semi-supervised learning. With spatiotemporal reasoning, our method is able to exploit large amount of unlabeled data to boost original performance. However, there is still a large gap between the performance of semi-supervised and supervised training (last two rows).

Furthermore, we show the performance of the student on nuScenes after each training iteration (Table \ref{tab:iterative}). By iteratively refining the teacher and the student, and gradually adding a small batch of unlabeled data, the performance of the student keeps improving.

\textbf{Ablations: }
In order to determine the contributions of each component of our method, we evaluate five different versions of our method changing one component of our method at a time:
\begin{itemize}
    \item Ours (Tracking~\cite{Weng2020_AB3DMOT}): Instead of using our proposed temporal graph neural networks, we adopt a Kalman Filter based 3D multi-object tracker~\cite{Weng2020_AB3DMOT} to reason about the spatial-temporal information. We use the average detection score of each trajctory as confident score, and the most confident tracks are then used as pseudo labels to supervise the student network.
    \item Ours (Flicker~\cite{jin2018unsupervised}): Inspired by Flicker~\cite{jin2018unsupervised}, we propose to decrease the confidence of detections that are isolated in time, \textit{i.e.} that have no associated preceding or following detections, and we increase the confidences of detections that are near to high-scoring detections in adjacent frames.
    The most confident tracks are then used as pseudo labels to supervise the student network. 
    \item Ours (-Augmentations): We train the temporal GNN of the teacher without data augmentation, the node features and edge features remain unchanged. This ablation shows the value of data augmentation to avoid overfitting.
    \item Ours (-Gradually): Instead of training with more unlabeled samples gradually (adding a small batch of unlabeled data at each iteration), we use all unlabeled data at once for semi-supervised training.
    \item Ours (-Uncertainty): Our method trained without uncertainty weighting, where all pseudo labels are equally weighted in Eq.~\ref{eq:clss_loss}) during semi-supervised training. This ablation shows the value of uncertainty-aware training.
    \item Ours (-Iterative): Instead of refining the teacher and student iteratively, we leave the teacher network unchanged, while the student is finetuned with a  mix of labeled and unlabeled data.
\end{itemize}

% \dave{Additional ablations that we should do:}
% \begin{itemize}
%     \item Ours (No augmentations): No data augmentations (since we emphasize this as a novel component of our method
%     \item Ideally we should remove each data augmentation one-by-one
%     \item Ours: No uncertainty weighting (remove uncertainty weighting when training the student)
%     \item Ours: No isotonic regression
%     \item Ours: No iterative training
% \end{itemize}

% \dave{The naming of the methods in the table should match the naming above}

% \dave{You can combine the results and ablations into a single table; maybe put a line in the table between the methods that are baselines and the methods that are ablations}

\setlength{\tabcolsep}{6pt} %% default is 6pt
\begin{table}[!tbp]
% 	\vspace{-0.3cm}
	\caption{Performance of student after each training iteration on nuScenes. By iteratively refining the teacher and the student, and gradually adding a small batch of unlabeled data, the performance of the student keeps improving.}
% 	\vspace{-0.3cm}
	\fontsize{8}{6}\selectfont
	\begin{center}
	   % \resizebox{width=0.5\textwidth}{!}{
			\begin{tabular}{m{1.2cm}<{\centering}||m{2.8cm}<{\centering}|m{2.8cm}<{\centering}}
				\toprule
				\midrule
				& Unchanged Teacher & Refined Teacher\\
				\midrule 
 
				Iter-0	   &  23.17 & 23.17\\
				Iter-1    &  28.97 & 35.61\\
				Iter-2    &  \textbf{30.44} & \textbf{36.46}\\
				\bottomrule
			\end{tabular}
% 			}
	\end{center}
	\vspace{-0.6cm}
	\label{tab:iterative}
\end{table}

\setlength{\tabcolsep}{6pt} %% default is 6pt
\begin{table*}[!tbp]
% 	\vspace{-0.3cm}
	\caption{Comparison of Detection Performance (\%) on nuScenes Dataset}
% 	\vspace{-0.3cm}
	\fontsize{10}{8}\selectfont
	\begin{center}
		\resizebox{\textwidth}{!}{
			\begin{tabular}{m{3.2cm}<{\centering}||m{1.8cm}<{\centering}|m{1.8cm}<{\centering}|m{1.8cm}<{\centering}| m{1.8cm}<{\centering}| m{1.8cm}<{\centering}|m{1.8cm}<{\centering}|m{1.8cm}<{\centering}|m{1.8cm}<{\centering}|m{1.8cm}<{\centering}|m{1.8cm}<{\centering}|m{1.8cm}<{\centering}}
				\toprule
				\midrule
				Method & mAP & Car & Truck & Bus & Trailer & CV & Pedes & Motor & Bicycle & TC & Barrier\\
				\midrule 
                
                Student (w/o Semi)	 &  23.17 & 62.74 & 17.80 & 15.93 & 0.00 & 0.17 & 60.04 & 23.55 & 5.57 & 22.21 & 22.19\\
				Gradual Semi	 & 25.58 & 58.09 & 19.54 & 17.26 & 3.83 & 0.09 & 64.73 & 19.2 & 5.31 & 36.88 & 30.07\\ 
				SESS                     & 27.45 & 61.77 & 21.17 & 19.03 & 6.12 & 0.25 & 63.32 & 24.39 & 6.98 & 40.13 & 31.33\\
				Ours	                 & \textbf{36.46} & \textbf{74.94} & \textbf{29.86} & \textbf{31.32} & \textbf{10.14} & \textbf{3.76} & \textbf{75.64} & \textbf{31.91} & \textbf{12.29} & \textbf{49.09} & \textbf{45.61}\\
				\midrule 
				Supervised Training      & 41.22 & 75.98 & 38.00 & 39.56 & 16.77 & 10.37 & 79.20 & 39.95 & 18.88 & 50.54 & 42.95\\
				\bottomrule
			\end{tabular}
		}
	\end{center}
	\vspace{-0.6cm}
	\label{tab:nuScenes}
\end{table*}

% \setlength{\tabcolsep}{2pt} %% default is 6pt
% \begin{table*}[!tbp]
% % 	\vspace{-0.3cm}
% 	\caption{Ablation (\%) on H3D Dataset}
% % 	\vspace{-0.3cm}
% 	\fontsize{10}{8}\selectfont
% 	\begin{center}
% 		\resizebox{\textwidth}{!}{
% 			\begin{tabular}{m{3.2cm}<{\centering}||m{1.8cm}<{\centering}|m{1.8cm}<{\centering}|m{1.8cm}<{\centering}| m{1.8cm}<{\centering}| m{1.8cm}<{\centering}|m{1.8cm}<{\centering}|m{1.8cm}<{\centering}|m{1.8cm}<{\centering}|m{1.8cm}<{\centering}}
% 				\toprule
% 				\midrule
% 				Method & mAP & Car & Ped & Other vehicle & Truck & Bus & Motorcyclist & Cyclist & Animal\\
% 				\midrule 
 
% 				Ours Tracking	 & 30.98 & & & & & & & &\\
% 				Ours Flicker    &  33.85 & 54.22 & 62.01 & 10.83 & 33.81 & 11.67 & 30.10 & 61.10 & 7.09\\
% 				Ours Add all data once  &  38.22 & 56.31 & 62.52 & 14.59 & 30.79 & 13.72 & 29.24 & 60.39 & 9.8\\
% 				\bottomrule
% 			\end{tabular}
% 		}
% 	\end{center}
% 	\vspace{-0.6cm}
% 	\label{tab:nuScenes}
% \end{table*}

\setlength{\tabcolsep}{6pt} %% default is 6pt
\begin{table*}[!tbp]
% 	\vspace{-0.3cm}
	\caption{Comparison of Detection Performance (\%) on H3D Dataset}
% 	\vspace{-0.3cm}
	\fontsize{10}{8}\selectfont
	\begin{center}
		\resizebox{\textwidth}{!}{
			\begin{tabular}{m{3.2cm}<{\centering}||m{1.8cm}<{\centering}|m{1.8cm}<{\centering}|m{1.8cm}<{\centering}| m{1.8cm}<{\centering}| m{1.8cm}<{\centering}|m{1.8cm}<{\centering}|m{1.8cm}<{\centering}|m{1.8cm}<{\centering}|m{1.8cm}<{\centering}}
				\toprule
				\midrule
				Method & mAP & Car & Ped & Other vehicle & Truck & Bus & Motorcyclist & Cyclist & Animal\\
				\midrule 
 
				Student (w/o Semi)	 &  31.02 & 55.77 & 63.4 & 5.26 & 30.23 & 12.16 & 22.6 & 55.68 & 3.03\\
				Gradual Semi    &  35.23 & \textbf{56.25} & 63.03 & \textbf{16.56} & 35.39 & 14.49 & 23.72 & 63.28 & 9.09\\
				SESS                     &  34.11 & 55.91 & \textbf{63.12} & 14.43 & 33.37 & 13.62 & 24.87 & 59.42 & 8.14\\
				Ours	                 &  \textbf{38.99} &	56.22 & 63.01 & 11.83 & \textbf{35.81} & \textbf{14.67} & \textbf{27.26} & \textbf{64.10} & \textbf{9.09}\\
				\midrule 
				Supervised Training      &  45.56 &	61.27 & 55.76 & 19.1 &  41.56 & 19.51 & 49.93 & 71.82 & 2.46\\
				\bottomrule
			\end{tabular}
		}
	\end{center}
	\vspace{-0.6cm}
	\label{tab:h3d}
\end{table*}

\setlength{\tabcolsep}{6pt} %% default is 6pt
\begin{table*}[!tbp]
% 	\vspace{-0.3cm}
	\caption{Ablation Study on nuScenes Dataset}
% 	\vspace{-0.3cm}
	\fontsize{12}{10}\selectfont
	\begin{center}
		\resizebox{\textwidth}{!}{
		\begin{tabular}{m{4.8cm}<{\centering}||m{1.8cm}<{\centering}|m{1.8cm}<{\centering}|m{1.8cm}<{\centering}| m{1.8cm}<{\centering}| m{1.8cm}<{\centering}|m{1.8cm}<{\centering}|m{1.8cm}<{\centering}|m{1.8cm}<{\centering}|m{1.8cm}<{\centering}|m{1.8cm}<{\centering}|m{1.8cm}<{\centering}}
				\toprule
				\midrule
				Method & mAP & Car & Truck & Bus & Trailer & CV & Pedes & Motor & Bicycle & TC & Barrier\\
				\midrule 
				
				Ours (Tracking)	    & 29.69 & 65.43 & 23.14 & 21.56 & 7.65 & 3.14 & 64.23 & 27.89 & 7.32 & 42.01 & 34.52\\ 
				Ours (Flicker)             & 34.97 & 72.94 & 26.86 & 30.32 & \textbf{11.21} & \textbf{4.02} & 72.35 & 30.27 & 10.11 & 48.79 & 42.82\\
				Ours (-Data Augmentation)   & 33.30 & 70.51 & 23.03 & 30.12 & 8.34 & 3.21 & 71.98 & 30.32 & 10.11 & 43.07 & 42.32\\
				Ours (-Gradually Training)           & 32.20 & 64.73 & 28.09 & 27.34 & 8.75 & 3.18 & 71.73 & 28.11 & 10.01 & 45.41 & 34.67\\
				Ours (-Uncertainty)         & 30.61 & 62.34 & 26.51 & 15.01 & 4.32 & 3.61 & 70.21 & 29.97 & 9.81 & 44.03 & 40.32\\
				Ours (-Iterative)         & 30.44 & 61.34 & 28.51 & 18.01 & 7.32 & 3.87 & 68.32 & 27.92 & 7.56 & 43.23 & 38.32\\
				Ours	                    & \textbf{36.46} & \textbf{74.94} & \textbf{29.86} & \textbf{31.32} & 10.14 & 3.76 & \textbf{75.64} & \textbf{31.91} & \textbf{12.29} & \textbf{49.09} & \textbf{45.61}\\
				\bottomrule
			\end{tabular}
		}
	\end{center}
	\vspace{-0.6cm}
	\label{tab:ablation}
\end{table*}

\begin{figure}[t!]
\begin{center}
    % \centering
    \includegraphics[width=0.48\textwidth]{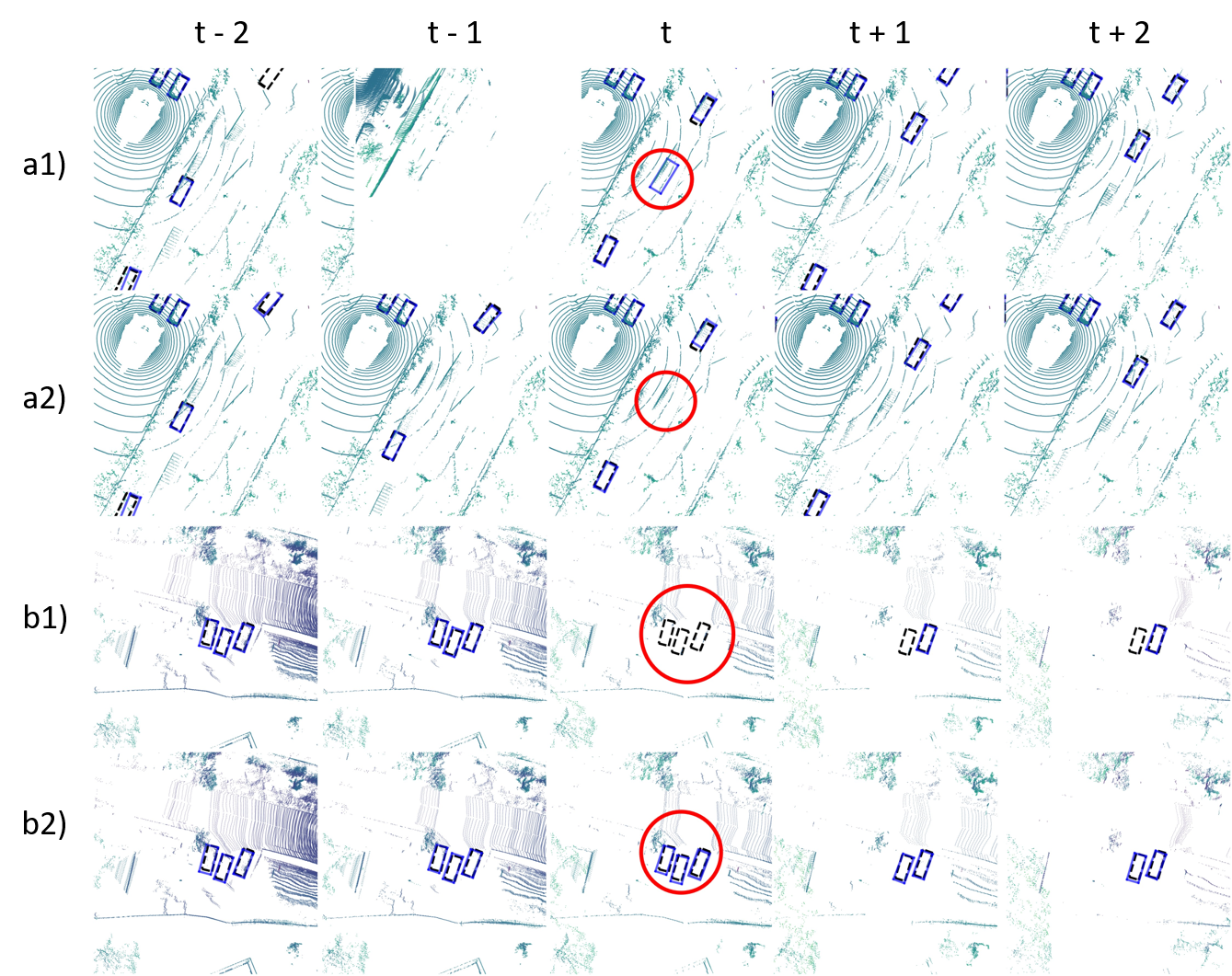}
    \caption{Qualitative Results: Ground truth detections are shown as black dashed boxes. Predicted detections are shown as blue boxes. We filter the detections with confidence lower than 0.5. The a1) and b1) show the detection results from initial student. The a2) and b2) rows show the result of our method. a2) shows our method successfully corrects false positive detections are near to high-scoring detections in adjacent frames (red circles). b2) shows our method corrects false negative detections that are isolated in time (red circles).}
     \vspace{-0.6cm}
    \label{fig:flicker}
\end{center}
% \vspace{-0.3cm}
\end{figure}
% \vspace{-0.6cm}

\begin{figure}[t!]
\begin{center}
    % \centering
    \includegraphics[width=0.4\textwidth]{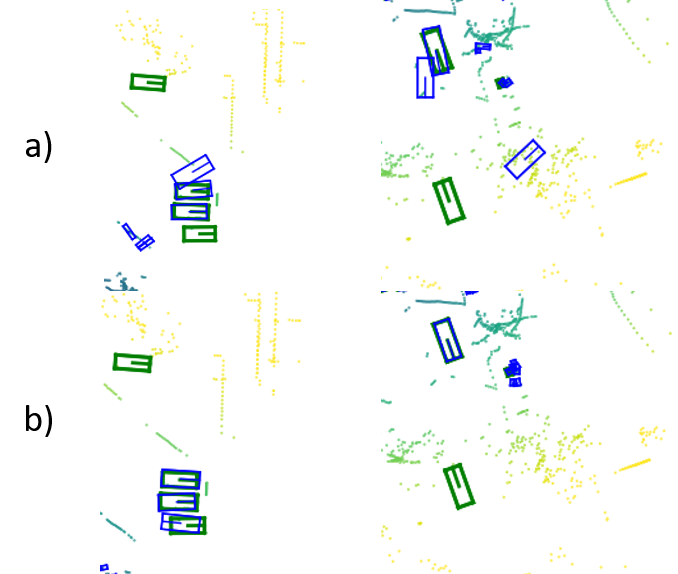}
    \caption{Qualitative Results (Semi-supervised): Ground truth detections are shown as green boxes. Predicted detections are shown as blue boxes. We filter the detections with confidence lower than 0.5. a) shows the results from the initial student. b) shows the result from the learned student. After training with unlabeled data via semi-supervised training, our method improves the detection results.}
     \vspace{-0.6cm}
    \label{fig:semi}
\end{center}
% \vspace{-0.3cm}
\end{figure}
% \vspace{-0.6cm}

According to the ablation study, we show that each component contributes to the final improvement of the performance. Comparing the first and last row of Table~\ref{tab:ablation}, we prove using the tracking method instead the GNN model can get worse performance, because tracking method updates not only the confident score but also the position of each detected object. These object position predicted by Kalman Filter introduces new errors to the pseudo label, which are then  learned by the student. Our (Flicker) indicates the GNN network of student can more efficient extract features and understand the relation among the objects from the point cloud video. Our (-Augmentations) shows the importance of graph augmentation for temporal GNN (comparing the third and last row of Table~\ref{tab:ablation}). The temporal GNN can easily overfit to the limited amount of labeled data without graph augmentation, and thus hurts the performance of the teacher and student. Our (-Gradually) indicates the importance of gradually training (comparing the fourth and last row of Table~\ref{tab:ablation}). Since the initial student is trained with limited label data, the model can easily get negative effect by large amount of pseudo labels with high uncertainty. Our (-Uncertainty) shows that even with spatiotemporal reasoning, the teacher's prediction is still noisy. Thus using uncertainty to weight pseudo labels significantly reduce the affect of unreliable labels and thus can help the student to learn more from good samples. Our (-Iterative) shows that keeping the teacher updated is another key component to boost the detection performance as a better teacher can generate more reliable pseudo labels. 

\subsection{Qualitative Analysis}
In this section, we qualitatively analyze our method. Figure~\ref{fig:flicker} a1) and a2) visualizes examples where our method corrects false positive detections made by the original teacher. a1) shows the detection from initial student, where in frame $t$, ground-truth objects were failed to be detected. However, the detector is able to detect these objects in most of the previous two and next two frames. These are examples of ``false positive flickers", where a detection suddenly appears in one frame and disappears in the next. Since the GNN of student aggregates information from neighboring frames, it reasons that it is unlikely the object exists in frame $t$. Our method a2) successfully removes these false detections as shown in the second and fourth rows.

Similarly, Figure~\ref{fig:flicker} b1) and b2) visualizes examples where the learned student corrects the initial teacher's false negative. In the b1), the teacher falsely detects an object in frame $t$. However, most of the neighboring nodes do not have detections corresponding to this object. These are examples of ``false negative flickers" where an object is predicted to disappear in one frame and immediately reappear in the next frame. Our method takes temporal information into account, and reasons that because the neighboring frames have positive detections at the same location, it is highly likely that there should be a detection in frame $t$. The b2) shows that our method increases the confidence of the flickered detections and correctly detects the missed objects.

Figure~\ref{fig:semi} shows the comparison between a) the initial teacher and b) the student trained by our proposed semi-supervised approach in two different scenarios. The detection from the pre-trained model with limited data shows lots of false positive and false negation. After feeding into more unlabeled data, the student network produces a more accurate detection results by learning both positive and negative sample with low uncertainty.

Overall, our approach combines the temporal information which reduces the uncertainty of pseudo labels and semi-supervised training pipeline which obtains the benefits from larger set of data to improve the generalization and robustness of the model.

%% file: sections/5-conclusion.tex
\section{Conclusion}

In conclusion, we propose to leverage large amounts of unlabeled point cloud videos by semi-supervised learning of 3D object detectors via temporal graph neural networks.  It does not require a large amount of strong labels that are often difficult to obtain. We show that teacher with temporal GNN can generate more accurate pseudo labels to train the student. By incorporating uncertainty-aware semi-supervised training, gradual semi-supervised training, and iterative refinement, our method achieves state-of-the-art detection performance on the challenging nuScenes~\cite{caesar2020nuscenes} and H3D~\cite{360LiDARTracking_ICRA_2019} benchmarks, compared to baselines trained on the same amount of labeled data. We hope that our work points toward moving away from spending excessive efforts annotating labeled data and instead redirecting them to semi-supervised learning on large unlabeled dataset.

\minisection{Acknowledgements.} The authors would like to thank members of R-pad for fruitful discussion and detailed feedback on the manuscript. Carnegie Mellon Effort has been supported by the Honda Research Institute USA, NSF S\&AS Grant No. IIS-1849154.